\documentclass[sigconf]{acmart}
\AtBeginDocument{%
  }

\usepackage{microtype}
\usepackage{hyperref}
\usepackage{url}
\usepackage{booktabs}
\definecolor{darkblue}{rgb}{0, 0, 0.5}
\hypersetup{colorlinks=true, citecolor=darkblue, linkcolor=darkblue, urlcolor=darkblue}

\usepackage{array}
\usepackage{graphicx}
\usepackage{tabularx, caption, boldline}
\usepackage{cellspace}
\usepackage{algorithm}
\usepackage{algpseudocode}
\usepackage{amsmath}
\usepackage{amssymb}
\usepackage[noabbrev]{cleveref}
\usepackage{breqn}
\usepackage{adjustbox}
\usepackage{booktabs}
\usepackage{multirow}
\usepackage[nobiblatex]{xurl}
\usepackage{pgfplots}
\usepgfplotslibrary{groupplots}
\pgfplotsset{compat=1.17}
\usepackage{tikz}
\usepackage{subcaption}
\usepackage{caption}
\usepackage{svg}

\usepackage{tcolorbox}
\tcbuselibrary{breakable}
\usepackage{soul} 
\usepackage{xcolor} 
\definecolor{lightyellow}{rgb}{1,1,0.8}
\definecolor{lightred}{rgb}{1.0, 0.8, 0.8}
\definecolor{lightgreen}{rgb}{0.8,1,0.8}
\usepackage{framed}
\definecolor{promptbg}{RGB}{240,240,240}
\definecolor{promptframe}{RGB}{200,200,200}

\usepackage{caption}

\setcopyright{acmlicensed}
\copyrightyear{2018}
\acmYear{2018}
\acmDOI{XXXXXXX.XXXXXXX}

\acmConference[Conference acronym 'XX]{Make sure to enter the correct
  conference title from your rights confirmation emai}{June 03--05,
  2018}{Woodstock, NY}
\acmISBN{978-1-4503-XXXX-X/18/06}

\begin{document}

\title{Illuminating Blind Spots of Language Models with Targeted Agent-in-the-Loop Synthetic Data}

\author{Philip Lippmann \hspace{16pt} Matthijs T.J.~Spaan \hspace{16pt}  Jie Yang}
\affiliation{
  \institution{Delft University of Technology}
  \city{Delft}
  \country{The Netherlands}
}
\email{p.lippmann@tudelft.nl}



\begin{abstract}
Language models (LMs) have achieved impressive accuracy across a variety of tasks but remain vulnerable to high-confidence misclassifications, also referred to as unknown unknowns (UUs). 
These UUs cluster into blind spots in the feature space, leading to significant risks in high-stakes applications.
This is particularly relevant for smaller, lightweight LMs that are more susceptible to such errors.
While the identification of UUs has been extensively studied, their mitigation remains an open challenge, including how to use identified UUs to eliminate unseen blind spots. 
In this work, we propose a novel approach to address blind spot mitigation through the use of intelligent agents -- either humans or large LMs -- as teachers to characterize UU-type errors.
By leveraging the generalization capabilities of intelligent agents, we identify patterns in high-confidence misclassifications and use them to generate targeted synthetic samples to improve model robustness and reduce blind spots. 
We conduct an extensive evaluation of our method on three classification tasks and demonstrate its effectiveness in reducing the number of UUs, all while maintaining a similar level of accuracy.
We find that the effectiveness of human computation has a high ceiling but is highly dependent on familiarity with the underlying task.
Moreover, the cost gap between humans and LMs surpasses an order of magnitude, as LMs attain human-like generalization and generation performance while being more scalable.
\end{abstract}



\received{20 February 2007}
\received[revised]{12 March 2009}
\received[accepted]{5 June 2009}

\settopmatter{printacmref=false}

\maketitle

\section{Introduction}
\label{sec:intro}

\begin{figure*}[t]
    \centering
    \includegraphics[width=\textwidth]{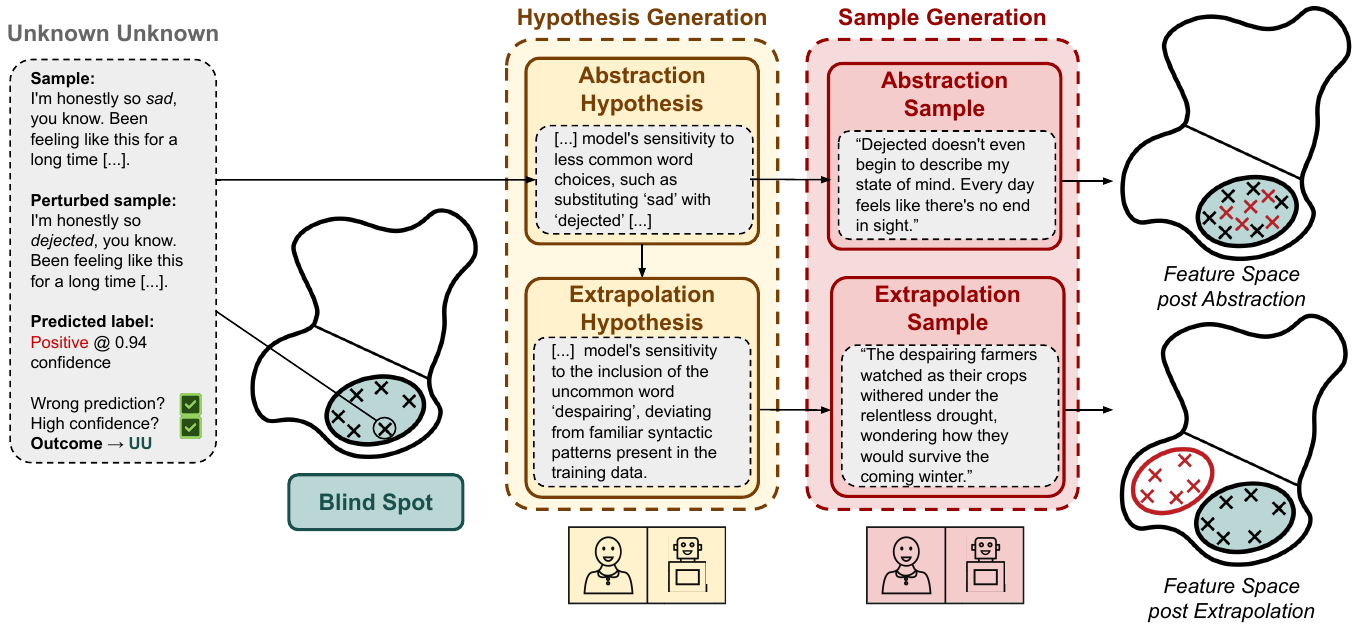}
    \caption{In a sentiment classification task, we begin with a UU resulting from a perturbation -- denoted by a cross in the feature space. 
    This UU is then used to generate an initial hypothesis via abstraction through human computation or an LM. 
    This abstraction hypothesis can then either by used to generate a synthetic samples that target the existing blind spot or to generate a new hypothesis via extrapolation, which in turn is then used to generate synthetic samples targeting an unseen blind spot.}
    \label{fig:generalization}
\end{figure*}

Language models (LMs) have achieved remarkable accuracy across a wide range of predictive tasks, but remain vulnerable to out-of-distribution data~\citep{Papernot2016adversarial, Wang2019adversarial, Brown2020gpt}. 
Small, lightweight LMs -- while easier to train and run on limited hardware, and therefore favored in domain-specific applications -- are especially prone to UUs due to their reduced robustness~\citep{wang2022measure, du2023robustness}.
Larger LMs, although generally more robust, require significant computational resources for both training and inference, limiting their usability~\citep{touvron2023llama2openfoundation}.
This vulnerability often leads to prediction errors, including in high-stakes applications such as suicide prevention~\citep{Large2017suicide} and criminal justice sentencing~\citep{Crawford2016agnostic}, where reliable and unbiased predictions are critical. 
A particularly challenging class of errors, referred to as \emph{unknown unknowns} (UUs), occurs when the model confidently misclassifies an input as the incorrect label~\citep{Attenberg2015machine}.
These UUs tend to \emph{cluster} into \emph{blind spots} in the feature space, areas where the model consistently produces high-confidence misclassifications due to biases in the training data~\citep{Lakkaraju2017bandit, Liu2020detection}.
On the left side of \cref{fig:generalization} we show an example of a mispredicted label at a high confidence, resulting in a UU, that forms part of a blind spot.

The identification of UUs and blind spots has been extensively studied~\citep{Attenberg2015machine, Bansal2018uu, Vandenhof2019hybrid, Liu2020detection}, including approaches involving human oversight to aid in detection~\citep{Cabrera2021errors, Han2021iterative}. 
\emph{Mitigating} blind spots -- especially how to move from identified blind spots to unseen ones -- remains an unresolved challenge.
Simple approaches to tackling only \emph{already discovered} blind spots, such as relabeling previously identified UUs and using them for additional training~\citep{Han2021iterative}, do not scale and fall short of ensuring a holistic reduction in blind spots.
Thus the only blind spots of the model that can be illuminated using such reactive approaches are those that correspond to seen data, with those that correspond to unseen data remaining out of reach.

In this paper, we introduce an agent-in-the-loop workflow that proactively mitigates blind spots of LMs by employing intelligent agents -- either humans or large LMs -- to \emph{characterize} blind spots and subsequently generate targeted synthetic data.
We pose that the key to mitigating these blind spots lies in the generalization abilities of the agent, allowing them to hypothesize patterns of discovered UUs and similarities between seen and unseen UUs using prior knowledge~\citep{Gluck2011generalization,Banich2010generalization,allaway2020general}. 
To this end, we guide agents to formulate these hypotheses in natural language, either describing the found blind spot consisting of discovered UUs (abstraction) or reasoning about undiscovered blind spots (extrapolation), as is shown in \cref{fig:generalization}.
Using these hypotheses, we guide agents toward the generation of synthetic samples targeted at blind spots, improving the robustness of LMs through subsequent retraining by reducing the number of high-confidence misclassifications without sacrificing overall predictive accuracy.
Our workflow is designed to flexibly integrate intelligence from both humans and LMs, with specific mechanisms to incorporate human computation or LMs. 
Additionally, the workflow can incorporate existing adversarial attack methods to proactively illuminate blind spots, further enhancing its adaptability and effectiveness.

Our workflow proves to be a viable means of distilling knowledge from intelligent agents to small LMs, making them more robust while maintaining their lightweight advantages.
Through our comprehensive experiments, we find that our method is capable of substantially reducing the number of high-confidence misclassifcations without decreasing accuracy.
On average, we are able to reduce the number of UUs by 19.08\%. 
Further, we show that for our method LMs are more effective overall than human agents, achieving a 22.37\% reduction in UUs compared to a 15.78\% reduction when using human-generated data.
Additionally, LM-generated data are far more economical, making them a more scalable solution for improving the robustness of small models. 
Finally, we observe that humans surpass LMs in certain tasks, particularly those that align more closely with human intuition due to their greater familiarity to participants.

In summary, the contributions of this paper are as follows:
\begin{itemize}
    \item A new workflow that utilizes the generalization capabilities of intelligent agents to mitigate blind spots, employing targeted synthetic data generation through an identify-characterize-generate approach.
    \item A comparative study on the efficacy of humans and LMs in applying our workflow to a variety of classification use cases and classification models, demonstrating the task-dependency of human contributions and scalability of LM-derived data. 
\end{itemize}

\section{Agent-in-the-Loop Targeted Data Generation}
\label{sec:method}
Our proposed approach to blind spot mitigation involves engaging a human or LM in three tasks: \emph{hypothesis generation via abstraction}, \emph{hypothesis generation via extrapolation}, and \emph{synthetic sample generation}. 
These tasks are designed to characterize and mitigate blind spots, ultimately reducing high-confidence misclassification. 
The workflow is schematically illustrated in~\cref{fig:generalization}.
The human computation component of our study is implemented through a survey study, the details of which are provided in~\cref{sec:appendixA}, while the equivalent LM prompts are given in~\cref{sec:appendixB}.

\subsection{Problem Formulation}
For UU discovery, let the dataset be $\mathcal{D} = \{(x_1, y_1), ..., (x_n, y_n)\}$, where $x$ is the original text sample and $y$ the original ground truth label.
Without having access to $y$, a predictive model $\theta$ is tasked with generating a label prediction $y_p = \theta(x)$ at a confidence $c \in [0,1]$. 
Formally, a UU occurs when (1) $\theta$ predicts the wrong label $y_p \neq y$ and (2) the prediction is made with high confidence $c \geq \tau$.

In this work, in addition to dealing with the blind spots that naturally occur in models as a result of training, we make use of adversarial UU discovery, where we increase the number of misclassifications by introducing perturbations. 
For this, a black-box adversarial perturbation model $G$ generates perturbed samples $\bar{x} = G(x)$, where $\bar{x} \neq x$. 
The model $\theta$ is then used to predict new labels $y'_i = \theta(\bar{x}_i)$ at a confidence $c$. 
The resulting perturbed dataset, denoted $\mathcal{P}$, consists of the new samples and predicted labels $(\bar{x}, y')$. 
If a perturbation occurs, there is an additional requirement for a misclassification to be considered a UU: (3) $\bar{x}$, regardless of its label indicated by $\theta$, maintains the same underlying true label $y$ as $x$ post perturbation.

Given a predictive model $\theta$ trained on a dataset $\mathcal{D}$, our objective is to mitigate UUs produced by $\theta$.
To systematically reduce high-confidence misclassification, we seek to identify patterns in discovered UUs and generate targeted synthetic data $\{x^s, y^s\}$ for a set of UUs, where $x^s$ is the synthetic label and $y^s$ represents the corresponding ground truth label for the synthetic sample.
This data is then used to further train $\theta$ and thus reduce the blind spots present.

\subsection{Generalization via Hypothesis Creation}
\label{sec:hyp}

For UU mitigation, we employ intelligent agents (humans or large LMs) to generalize from identified UUs to create hypotheses in natural language regarding the underlying causes of these UUs. 
As we use perturbations, such hypotheses are based on pairs of original and perturbed samples, $(x_i, y_i) \sim \mathcal{D}$ and $(\bar{x}_i, y_i') \sim \mathcal{P}$. 
Humans are adept at using sparse data to generalize~\citep{Lake2015human}, and this task exploits that capability by focusing on subsets of UUs. 
Each hypothesis describes the shared characteristics that explain why certain UUs occur and how these characteristics might generalize to other, unseen UUs.
The goal is not merely to explain individual failure cases but to construct hypotheses that address multiple UUs clustering together into a blind spot. 
In doing so, we can illuminate patterns within the feature space that the model is consistently misclassifying. 
To this end, we pursue two distinct but complementary strategies: abstraction and extrapolation.

\subsubsection{Abstraction} 
Abstraction involves generating a hypothesis on why a specific UU occurred that generalizes across a set of closely related UUs, revealing underlying patterns within a blind spot.
In this step, the intelligent agent is provided with an original sample $(x_i, y_i)$ and, if adversarial perturbations are used, its perturbed counterpart $(\bar{x}_{i}, y_{i}')$.
Then the agent is tasked with reasoning abstractly about the factors leading to this UU.
Specifically, we instruct them to consider whether these factors involve semantics, syntax, specific words, or something else in the samples that could be the cause of the high-confidence misclassification.
This is to guide the agent to identify what most likely contributes to the UU without prescribing rigid criteria, leaving room for creative thinking and allowing the agent to explore unforeseen or nuanced factors.
The hypothesis is in natural language and should generalize across other UUs that share these characteristics, expanding our understanding of the particular blind spot the UU corresponds to.
Compared to a mitigation approach that only makes use of a simple reactive relabeling of found UUs, our method comes with the additional advantage that it builds up a corpus of human-interpretable error reports on seen errors of the classification model.
An example of hypothesis generation via abstraction is shown in~\cref{fig:fullexample}.

\subsubsection{Extrapolation}
Extrapolation extends the process of hypothesis creation beyond trying to describe discovered blind spots, encouraging the agent to use existing hypotheses and sample pairs (used during abstraction) to uncover new blind spots. 
This task emphasizes extrapolation, asking the agent to hypothesize new failure modes -- also in natural language -- that differ from those previously identified. 
Extrapolative thinking has previously been shown to be a human strong suit~\cite{Bartlett1958thinking}.
By ensuring that the new hypotheses are dissimilar from those used for abstraction, we aim to discover new regions in the feature space where the model may be prone to high-confidence misclassification. 
To avoid the agent overextrapolating, we specifically instruct them to focus on the same topic but reason if a different possible factor from semantics, syntax, specific words could be at fault that was not mentioned in the abstraction hypothesis.
In this step, we present only human-generated hypotheses to human participants and vice versa. 
An example of extrapolation is shown in~\cref{fig:fullexample}.

\subsection{Synthetic Sample Generation}
Once hypotheses have been generated via abstraction or extrapolation, the agent is tasked with generating synthetic samples. 
These synthetic samples must align with the structure and context of the original dataset while reflecting the characteristics of the generated hypotheses. 
For instance, if the dataset consists of movie reviews, the synthetic samples should maintain the form and tone of movie review-related text.
The goal of this step is to create new data points that correspond to the blind spots identified during hypothesis generation. 
These synthetic samples are added to the training dataset, resulting in a dataset that is extended for each synthetic sample and its corresponding label $\mathcal{E} = \mathcal{D} \cup \{x^s_i, y^s_i\}$, where the label is provided by the agent. 
By incorporating these new samples into training, we aim to enhance the robustness of the predictive model $\theta$ by reducing its susceptibility to high-confidence misclassifications. 
The sample generation process is uniform, regardless of whether the hypothesis was obtained through abstraction or extrapolation. 
Humans generate samples based on human-created hypotheses, and LMs do the same for LM-generated hypotheses.
An example of this type of sample generation from human and LM agents for abstraction and extrapolation is shown in \cref{fig:fullexample}.

\begin{figure*}[t]
\centering
\begin{tcolorbox}[colback=gray!10!white, colframe=gray!50!black, width=\textwidth, title=Example of full workflow for a single UU sample using human computation and LMs on the IMDB Dataset, boxrule=0.5mm, arc=2mm, outer arc=2mm]

    \textbf{Original Sample:} \\
    Footage of busy highways, planes flying overhead, poor lightning effects, soldiers doing their thing, they're all used in a [[haphazard]] manner, sometimes repeatedly, and they often only have a loose connection to the story Wood is trying to tell. [...]

    \medskip

    \textbf{Perturbed Sample:} \\
    Footage of busy highways, planes flying overhead, poor lightning effects, soldiers doing their thing, they're all used in a [[thoughtless]] manner, sometimes repeatedly, and they often only have a loose connection to the story Wood is trying to tell. [...]

    \medskip
    \hrule
    \medskip
    
    \textbf{Human Abstraction Hypothesis:} \\
    ``Haphazard can be intentional, vs thoughtless is unplanned. It feels more of an unwillingness rather than a lack of ability.''

    \medskip

    \textbf{LM Abstraction Hypothesis:} \\
    ``The replacement of specific adjectives in the perturbed sample contributes to misclassification. The adjective "thoughtless" in the perturbed sample, may change the perception of the director's skills and use of stock footage. This alteration in sentiment could lead to a different classification outcome.''
    
    \medskip
    \hrule
    \medskip

    \textbf{Human Extrapolation Hypothesis:} \\
    ``The hypothesis could be that the words were potentially swapped or replaced, such as "bad" being replaced by "evil" which could lead to an incorrect prediction as it is similar but could have a slightly different meaning in some contexts.''

    \medskip

    \textbf{LM Extrapolation Hypothesis:} \\
    ``The use of the word "hypocrite" may have caused the misclassification. The word "hypocrite" implies that the person is saying one thing but doing another, which may be considered a negative trait. However, some people may not interpret the person in the sample as a hypocrite, leading to a difference in sentiment analysis. This difference in interpretation may have caused the misclassification by the algorithm.''

    \medskip
    \hrule
    \medskip

    \textbf{Human Abstraction Sample:} \\
    ``The scenes felt like they were slapped together with no care at all. It’s not that the director didn’t have the skill they just didn’t bother to try harder. Haphazard is at least creative but thoughtless just shows they didn’t wanna put the effort in.''

    \medskip

    \textbf{LM Abstraction Sample:} \\
    ``Skyward Dreams had potential, but the director's use of stock footage feels thoughtless. Scenes of bustling cities and planes flying overhead are inserted without any real purpose, disrupting the flow of the narrative. [...]''

    \medskip

    \textbf{Human Extrapolation Sample:} \\
    ``The CGI in was straight up evil. The way the effects looked completely ruined the immersion for me, and it felt like the creators didn’t even care about quality. I get that sometimes budget is an issue, but this was just on another level. [...]''

    \medskip

    \textbf{LM Extrapolation Sample:} \\
    ``The protagonist of The Final Betrayal is a true hypocrite. Throughout the film, he preaches loyalty and honesty to his friends, yet secretly manipulates and betrays them behind their backs. This hypocrisy is central to the film’s conflict, as the character's outward morality sharply contrasts with his deceitful actions. Despite this glaring flaw, some viewers may interpret his behavior as a survival tactic in a harsh world, rather than outright hypocrisy. [...]''

\end{tcolorbox}
\captionsetup{aboveskip=0pt}
\caption{Example of hypothesis generalization using \emph{abstraction} for the IMDB dataset. The abstraction is performed by a human or LLM based on original and perturbed samples.}
\label{fig:fullexample}
\end{figure*}

\begin{figure}[t]
    \centering
    \includegraphics[width=\linewidth]{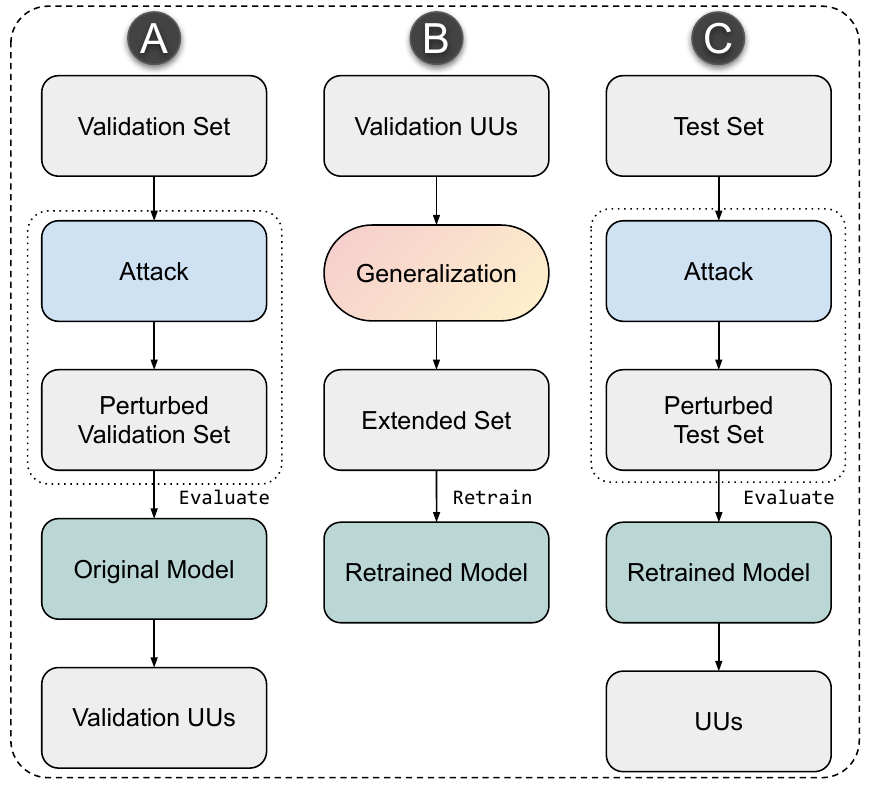}
    \caption{Workflow: (A) Obtain UUs from the validation set on the original finetuned model; (B) use UUs to extend the training data via generalization (\cref{fig:generalization}) and thus obtain a more robust model; (C) evaluate this retrained model. Adversarial perturbations in dotted box are optional.}
    \label{fig:workflow}
\end{figure}

\section{Experimental Setup}
\label{sec:experiment}
In this section, we present an overview of our experimental design. 
A schematic illustration of the workflow can be found in~\cref{fig:workflow}.
Here we first obtain our initial set of UUs of the finetuned classification model from the validation set. 
Following this, we characterize the blind spots corresponding to these UUs by making the intelligent agent perform generalization as described in \cref{sec:method}, culminating in new synthetic data that we use to retrain the model.
Finally, we evaluate this retrained model with respect to accuracy and UU count.
As a preliminary study, to verify that our method does indeed address blind spots, we successfully demonstrate that it is possible to artificially create blind spots by hand (\emph{i.e.}, ground truth blind spots) in a model and then illuminate these using our approach in \cref{sec:synthetic}.
In our main study, our experiments instead address mitigating both natural blind spots that occur during normal model training and those created by adversarial attacks. 
For this, we do not have access to the ground truth blind spots and as such just have indirect evidence that some blind spots are illuminated as the number of occurring UUs is decreased.

\subsection{Datasets, Models, and Perturbations}
To evaluate the generality and effectiveness of our approach, we select a diverse set of classification tasks, each representing varying levels of task complexity. 
Specifically, we focus on sentiment analysis (SA) using the IMDB dataset~\citep{Mass2011imbd}, semantic equivalence (SE) using the MRPC dataset~\citep{dolan2005mrpc}, and natural language inference (NLI) using the QNLI dataset~\citep{Rajpurkar2016squad}.
The statistics of the dataset for each task are shown in \cref{tab:datasets}.
For blind spot mitigation, we use the validation set to obtain our UUs that are then used to perform the hypotheses generalization. 
These hypotheses are then used in turn to generate synthetic samples and extend the training set, as shown in \cref{fig:workflow}. 
We limit the number of hypotheses derived from each of abstraction and extrapolation to 1\% of the training set size, leading to an additional 73, 500, and 2095 training samples after applying our method for MRPC, IMDB, and QNLI, respectively.
These values are treated as hyperparameters and are chosen to balance computational efficiency and effectiveness. 
We leave further optimization of this split between abstraction- and extrapolation-derived hypotheses to future work. 
We employ two classification models in our experiments, finetuned for each classification task: BERT (\texttt{bert-base-uncased})~\citep{Devlin2019bert} and Llama 2 (\texttt{llama-2-7b})~\citep{touvron2023llama2openfoundation}, selected for their contrasting architecture and size. 
We choose BERT for its known performance on sentence-level tasks and its low number of parameters, while Llama 2 was chosen for its larger (but still manageable) scale and capability in handling more complex language understanding tasks.
GPT-3.5 (\texttt{gpt-3.5-turbo-1106})~\citep{brown2020languagemodelsfewshotlearners} is incorporated as the teacher model to perform hypothesis and sample generation, as it is superior to both classification models that we use.

In a black-box setting, where we assume no access to the model’s internal parameters, we employ adversarial perturbation techniques to yield more UUs for our method to use. 
Note that while perturbations aid proactive discovery of blind spots, they are not strictly necessary to our overall approach.
Perturbations are generated using TextAttack~\citep{morris2020textattack}, specifically with TextFooler (TF)~\citep{Jin2020text} for word-level perturbations and DeepWordBug (DWB)~\citep{Ji2018DeepWordBug} for character-level perturbations. 
Using these two methods, we cover a wide spectrum of adversarial attack types, revealing additional blind spots.
We focus on perturbations that maintain semantic integrity, ensuring that the true underlying label remains consistent after perturbation. 
Manual inspection of 100 random perturbed samples revealed that none had a different underlying true label, affirming that our perturbations are faithful.

\subsection{Baseline}
As a baseline, we use a reactive relabeling approach based on the previous work by ~\citep{Han2021iterative}, where identified UUs are given a ground truth label, before being reintroduced to the classification model for additional training.
This method directly targets blind spots by adding these correctly labeled samples to the extended set. 
While ~\citep{Han2021iterative} performs this reintroduction in smaller, iterative batches to identify more UUs, we pool all relabeled UUs in a single batch, as we only concern ourselves with the mitigation of UUs and assume that we have knowledge of whether a sample is a UU or not post classification.
This is similar to how we perform the retraining for our method.
For a fair comparison, we apply this baseline approach with the same budgetary constraints as our proposed method, with new samples making up 2\% of the initial training set size.
We pose that our method, which uses hypotheses to synthesize new data, will outperform this method by uncovering additional failure modes not captured by relabeling alone.

\subsection{Implementation}
\label{sec:impl}
Following~\citet{Lakkaraju2017bandit}, we set the confidence threshold for determining high-confidence misclassifications to $\tau = 0.65$.
We use GPT-3.5 with a temperature setting of $T = 0.7$ and the default system message.
All BERT models were trained for 10 epochs, using a learning rate of $2 \times 10^{-4}$, and a batch size of 64.
We fine-tune all Llama 2 7B models using the Low-Rank Adaptation (LoRA) \citep{hu2021loralowrankadaptationlarge} method with the following configuration: a LoRA scaling factor of 16, dropout of 0.1, and rank $r=64$. The target modules are all linear layers in the model, and no bias adjustment is applied. 
The training is performed over 3 epochs, with a batch size of 8, and gradient accumulation set to 8 steps. 
We employ AdamW as optimizer. 
The learning rate is set to $2 \times 10^{-4}$ with a cosine learning rate schedule and a warmup ratio of 0.03. 
We apply a maximum gradient norm of 0.3 to ensure stability during training.
We use a weight decay of 0.001 to prevent overfitting.

The human computation component of our study is implemented through a survey study, the details of which are provided in~\cref{sec:appendixA}.
A key procedural difference between human and LM-based experiments is the number of examples provided.
The human participants receive two examples, while no examples are given to LMs (\emph{i.e.}, zero-shot). 
This design choice aims to minimize guidance for the LM since few-shot prompting tends to result in overly homogeneous samples, even when using higher temperature settings.
The LM prompts for the teacher model are given in~\cref{sec:appendixB}.
When prompting the teacher model, we always ask it to explicitly give its reasoning, which we find not only increases performance but also improves interpretability.
To ensure the quality of human-generated hypotheses and synthetic samples, we include attention checks~\citep{OPPENHEIMER2009attention} in each survey to eliminate inattentive or low-effort responses. 
For both human- and LLM-generated hypotheses and samples, we implement automated quality checks for this purpose. 
We do not focus on selecting the high-quality responses, but filter out bad-faith ones such as repeated or nonsensical submissions.
To be included, all text entries are required to meet a minimum character threshold ($\text{char}_{\text{min}} = 40$) to ensure sufficient content.
Additionally, we employed BERTScore~\citep{bert-score} to automatically evaluate the similarity of new samples against a reference set in the form of samples from the training set. 
If the similarity score falls below a threshold of $S_{min} = 0.5$, the entry is discarded.

\begin{table}[t]
\caption{Datasets used, including the task type, number of classes, and number of samples in each of the test, validation, and training sets. Note the split of the original IMDB test set into new validation and test sets.}
\centering
\begin{tabular}{lccccc}
\toprule
\multicolumn{1}{l}{\textbf{Dataset}} & \multicolumn{1}{l}{\textbf{Task}} & \multicolumn{1}{l}{\textbf{\#Classes}} & \multicolumn{1}{l}{\textbf{\#Train}} & \multicolumn{1}{l}{\textbf{\#Validation}} & \multicolumn{1}{l}{\textbf{\#Test}}\\ \midrule
MRPC   & SE & 2 & 3,668 & 408 & 1,725       \\
IMDB   & SA & 2 & 25,000 & 12,500 & 12,500        \\
QNLI   & NLI & 2 & 104,743 & 5,463 & 5,463          \\ 
\bottomrule
\end{tabular}
\label{tab:datasets}
\end{table}

\subsection{Evaluation Metrics}
We use two key metrics to assess the effectiveness of our approach and the comparative approach. 
These include the accuracy of the model on the test set and the number of UUs observed during evaluation. 
Accuracy provides a basic measure of model performance, while the UU count reflects the model's robustness and allows us to reason about the prevalence of blind spots.
Note that the accuracy we report is the accuracy of the model before any perturbations are applied, while the number of UUs is post perturbation.
Ideally, our goal is to maximize accuracy while minimizing the number of UUs. 
Our evaluation compares the performance of the original finetuned model with that of the models retrained on their respective extended dataset $\mathcal{E}$. 
This allows us to quantify the impact of our approach on mitigating blind spots and improving model robustness.

\begin{table*}[t]
\caption{Results of the blind spot study across datasets for BERT and Llama 2 7B as classification models. Here TF refers to the TextFooler perturbation method and DWB to DeepWordBug. An \(\uparrow\) indicates that a higher score is preferable, while \(\downarrow\) indicates that lower is better.}
\centering
\setlength{\tabcolsep}{5pt}
\begin{tabular}{clcccccccccc}
\toprule
& & \multicolumn{4}{c}{\texttt{BERT}} & & \multicolumn{4}{c}{\texttt{Llama 2 7B}} \\
\cmidrule(lr){3-6} \cmidrule(lr){8-11}
& & \multicolumn{2}{c}{\textbf{TF}} & \multicolumn{2}{c}{\textbf{DWB}} & & \multicolumn{2}{c}{\textbf{TF}} & \multicolumn{2}{c}{\textbf{DWB}} \\
\cmidrule(lr){3-4} \cmidrule(lr){5-6} \cmidrule(lr){8-9} \cmidrule(lr){10-11}
& & Acc (\%) \(\uparrow\) & UUs (\#) \(\downarrow\) & Acc (\%) \(\uparrow\) & UUs (\#) \(\downarrow\) & & Acc (\%) \(\uparrow\) & UUs (\#) \(\downarrow\) & Acc (\%) \(\uparrow\) & UUs (\#) \(\downarrow\) \\
\midrule
\multirow{4}{*}{\rotatebox[origin=c]{90}{\textbf{MRPC}}}
 & Original Model & 82.38 & 952 & 82.38 & 936 & & 90.84 & 301 & 90.66 & 293  \\
 & Relabelling Baseline & 82.49 & 911 & 82.55 & 898 & & 90.61 & 277 & 90.73 & 268 \\
 & Our Method w/GPT-3.5 & 81.57 & 851 & 82.23 & 882 & & 89.86 & 149 & 89.73 & 164 \\
 & Our Method w/Humans & 81.58 & 418 & 82.10 & 802 & & 90.20 & 144 & 89.91 & 140 \\
\midrule
\multirow{4}{*}{\rotatebox[origin=c]{90}{\textbf{IMDB}}}
 & Original Model & 94.84 & 1882 & 95.40 & 1682 & & 95.20 & 892 & 95.33 & 810 \\
 & Relabelling Baseline & 93.94 & 1732 & 94.26 & 1621 & & 94.86 & 781 & 95.10 & 742 \\
 & Our Method w/GPT-3.5 & 95.40 & 1241 & 94.41 & 1448 & & 94.96 & 604 & 95.13 & 689 \\
 & Our Method w/Humans & 94.43 & 1518 & 95.74 & 1412 & & 94.67 & 658 & 94.90 & 702 \\
\midrule
\multirow{4}{*}{\rotatebox[origin=c]{90}{\textbf{QNLI}}} 
 & Original Model & 89.88 & 1923 & 89.88 & 2597 & & 90.08 & 879 & 90.72 & 952 \\
 & Relabelling Baseline & 88.24 & 1796 & 88.98 & 1907 & & 89.90 & 856 & 90.60 & 929 \\
 & Our Method w/GPT-3.5 & 89.31 & 1536 & 89.21 & 1746 & & 89.58 & 741 & 90.10 & 890 \\
 & Our Method w/Humans & 89.42 & 2028 & 89.38 & 2325 & &  89.16 & 857 & 89.73 & 924 \\
\bottomrule
\end{tabular}
\label{tab:gptnaturalresults}
\end{table*}

\section{Results}
\label{sec:user}
\label{sec:natural}
In this section, we report the experimental results on the effectiveness of our proposed method in reducing blind spots across the classification tasks.
The results of our methods configured with human- and LM-generated data as well as those of the baselines are shown in~\cref{tab:gptnaturalresults}. 
Additionally, we compare human-generated samples to those produced by LMs in terms of effectiveness, scalability, and ease of use.

\subsection{Impact of Synthetic Samples}

\subsubsection{Observation 1: Our approach leads to a significant and consistent UU reduction across tasks}
As part of our evaluation, we find that our method successfully reduces UUs, with a maximum reduction of 56.09\% when using human computation on the BERT model with TF for the MRPC task.
On average, across perturbation methods and classification models, our method with LM-based data generation reduced UUs by 22.37\%, while human-based data generation led to a reduction of 15.78\%. 
Similarly, regardless of what type of agent generates the data, our method achieves a reduction in UUs of 35.77\%, 21.46\%, and 13.03\% for MRPC, IMDB, and QNLI, respectively.
These results highlight the strengths of using agent-generated samples, with large LMs as a teacher model generally offering more consistent reductions in UUs, though there are difference between tasks. 
The only configuration where our method does not reduce UUs is the BERT model on the QNLI dataset, where human-based retraining with TF actually increases UUs by 5.46\%. 
We elaborate on this in observation 3.

\subsubsection{Observation 2: Relabeling of UU samples is effective but not as impactful}
Simply relabeling UU samples from the validation set and reintroducing them as the extended set leads to a decrease in the number of UUs, albeit a more modest one compared to our method.
Relabeling achieves a consistent decrease in UUs across tasks of 11.10\% and 7.08\% on average for BERT and Llama 2, respectively, compared to an average decrease of 14.10\% and 17.45\% for our method when using humans and 22.74\% and 22.00\% when using LMs.
This confirms that only reactive illumination of blind spots using seen data is less effective than our method, regardless of agent type, as the characterization and subsequent extrapolation we employ results in a more significant reduction in UUs.
While the average decrease is lower, the relabeling method is very consistent across tasks, as it is not dependent on an agent grasping the task and delivering high quality data.
Additionally, it is very cost effective as no human computation or LM querying is necessary.
The obvious limitation of this approach is that it only scales to blind spots that have been discovered and therefore has very little transfer learning potential, as it is unlikely that the found UUs with generalize to unseen UUs.

\subsubsection{Observation 3: Human performance is very task dependent} 
We find that human-generated samples may outperform LMs in tasks that align with human intuition. 
For tasks such as SE and SA -- which are more intuitive to humans compared to NLI, as they more closely resemble everyday tasks -- human performance tends to be better, yielding more significant reductions in UUs.
In particular, on the MRPC dataset we see a greater reduction in UUs using human-generated data, 35.38\% and 52.19\% on BERT and Llama 2, respectively, when compared to when using LM-generated hypotheses and samples 8.21\% and 47.31\%. 
In less intuitive tasks such as NLI, humans can generate data of poor quality, leading to a reduction in model robustness, which may even result in an increase in UUs.
When analyzing participants' responses for QNLI, we find that several participants did not fully grasp the natural language inference task, which was not the case for SE and SA.
Note that these are not purposefully low-effort responses and are therefore not filtered out as described in \cref{sec:impl}.
This shows that irrespective of classification model, there is a task-specific advantage of human computation compared to LM teacher models when there exists a higher degree of familiarity with the task and vice versa.
Although LMs provide samples of acceptable quality consistently, rare but high-quality human responses, such as a crowdworker correctly identifying that changing the date ``June 15'' to ``John 15'' referenced a Bible verse -- an insight that the LM missed -- can significantly reduce UUs and thus be more impactful. 
This suggests that while human-generated responses can have a higher ceiling in certain contexts, LMs deliver more consistent results overall as just a few human participants' incorrect responses can reduce the effectiveness of our method.

\subsubsection{Observation 4: Accuracy does not decrease despite improved robustness}
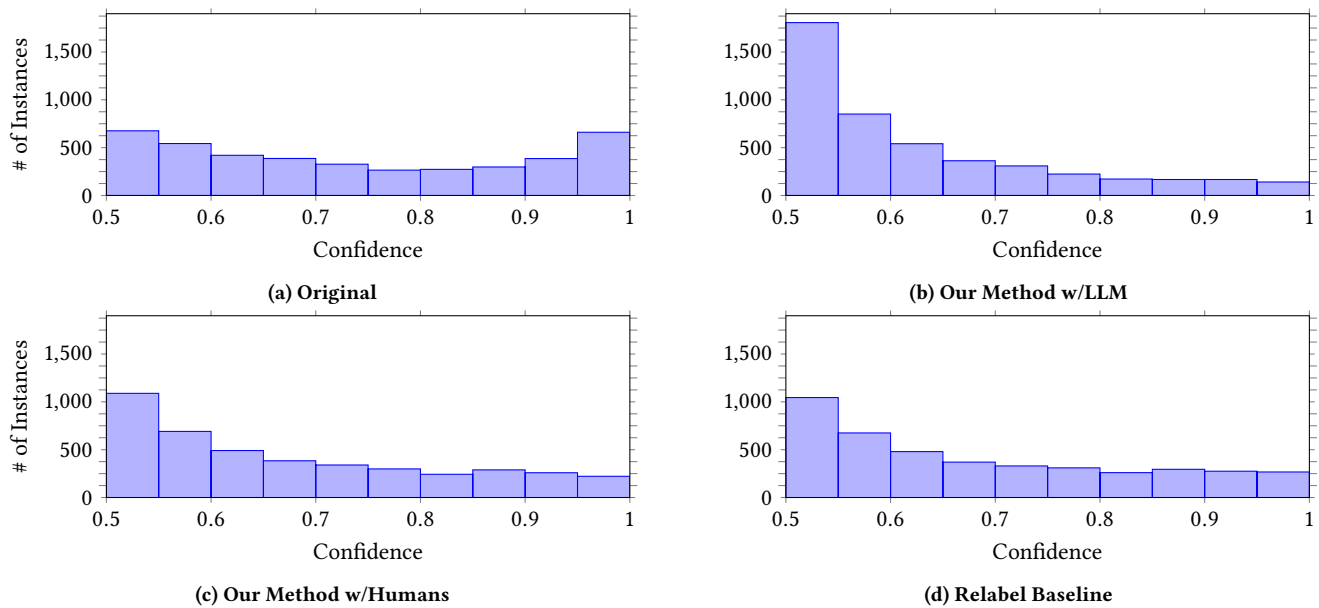
\begin{figure*}[t]
    \centering
    
    \begin{subfigure}[b]{0.48\textwidth}
        \centering
        \begin{tikzpicture}
            \begin{axis}[
                width=\linewidth,
                height=4cm,
                xmin=0.5, xmax=1,
                ymin=0, ymax=1900,
                xtick={0.5,0.6,0.7,0.8,0.9,1},
                xticklabel style={rotate=0},
                minor y tick num = 3,
                area style,
                xlabel={Confidence},
                ylabel={\# of Instances},
                axis line style={-},
                tick align=outside,
                tick style={line width=0.4pt},
                major tick length=2pt,
                every axis plot/.append style={fill=cyan!70},
                ]
                \addplot+[ybar interval] plot coordinates { 
                    (0.5, 676)
                    (0.55, 543)
                    (0.6, 420)
                    (0.65, 387)
                    (0.7, 327)
                    (0.75, 265)
                    (0.8, 273)
                    (0.85, 298)
                    (0.9, 386)
                    (0.95, 661)
                    (1, 0)
                };
            \end{axis}
        \end{tikzpicture}        
        \caption{Original}
        \label{subfig:hist1}
    \end{subfigure}
    \hfill
    \begin{subfigure}[b]{0.48\textwidth}
        \centering
        \begin{tikzpicture}
            \begin{axis}[
                width=\linewidth,
                height=4cm,
                xmin=0.5, xmax=1,
                ymin=0, ymax=1900,
                xtick={0.5,0.6,0.7,0.8,0.9,1},
                xticklabel style={rotate=0},
                minor y tick num = 3,
                area style,
                xlabel={Confidence},
                axis line style={-},
                tick align=outside,
                tick style={line width=0.4pt},
                major tick length=2pt,
                every axis plot/.append style={fill=cyan!70},
                ]
                \addplot+[ybar interval] plot coordinates { 
                    (0.5, 1807)
                    (0.55, 851)
                    (0.6, 541)
                    (0.65, 363)
                    (0.7, 309)
                    (0.75, 224)
                    (0.8, 172)
                    (0.85, 167)
                    (0.9, 167)
                    (0.95, 141)
                    (1, 0)
                };
            \end{axis}
        \end{tikzpicture}        
        \caption{Our Method w/LLM}
        \label{subfig:hist2}
    \end{subfigure}
    
    \begin{subfigure}[b]{0.48\textwidth}
        \centering
        \begin{tikzpicture}
            \begin{axis}[
                width=\linewidth,
                height=4cm,
                xmin=0.5, xmax=1,
                ymin=0, ymax=1900,
                xtick={0.5,0.6,0.7,0.8,0.9,1},
                xticklabel style={rotate=0},
                minor y tick num = 3,
                area style,
                xlabel={Confidence},
                ylabel={\# of Instances},
                axis line style={-},
                tick align=outside,
                tick style={line width=0.4pt},
                major tick length=2pt,
                every axis plot/.append style={fill=cyan!70},
                ]
                \addplot+[ybar interval] plot coordinates { 
                    (0.5, 1089)
                    (0.55, 691)
                    (0.6, 491)
                    (0.65, 384)
                    (0.7, 340)
                    (0.75, 299)
                    (0.8, 243)
                    (0.85, 289)
                    (0.9, 259)
                    (0.95, 222)
                    (1, 0)
                };
            \end{axis}
        \end{tikzpicture}        
        \caption{Our Method w/Humans}
        \label{subfig:hist3}
    \end{subfigure}
    \hfill
    \begin{subfigure}[b]{0.48\textwidth}
        \centering
        \begin{tikzpicture}
            \begin{axis}[
                width=\linewidth,
                height=4cm,
                xmin=0.5, xmax=1,
                ymin=0, ymax=1900,
                xtick={0.5,0.6,0.7,0.8,0.9,1},
                xticklabel style={rotate=0},
                minor y tick num = 3,
                area style,
                xlabel={Confidence},
                axis line style={-},
                tick align=outside,
                tick style={line width=0.4pt},
                major tick length=2pt,
                every axis plot/.append style={fill=cyan!70},
                ]
                \addplot+[ybar interval] plot coordinates { 
                    (0.5, 1045)
                    (0.55, 675)
                    (0.6, 480)
                    (0.65, 370)
                    (0.7, 330)
                    (0.75, 310)
                    (0.8, 260)
                    (0.85, 295)
                    (0.9, 275)
                    (0.95, 267)
                    (1, 0)
                };
            \end{axis}
        \end{tikzpicture}        
        \caption{Relabel Baseline}
        \label{subfig:hist4}
    \end{subfigure}
    \caption{Plots of prediction confidence per misclassified sample for BERT on QNLI dataset when using TF as a perturbation technique, showing the distribution across confidence bins. The distribution of the prediction confidences is altered by the retraining, regardless of how it was performed. Our method is able to lower the number of high-confidence classifications, especially those at the highest of confidences, improving model calibration.}
    \label{fig:four-histograms}
\end{figure*}

In terms of accuracy, extending the training set with human- or LM-generated data did not have a significant effect.
Across tasks, accuracy fluctuations of the models with extended training sets remain within $\pm$1\% compared to the original models.
This contrasts with previous findings that improvements in robustness often come at the expense of accuracy~\citep{tsipras2018robustness}.
To illustrate the impact of retraining on accuracy, we visualize prediction confidences across misclassified samples post perturbation for a selected dataset and perturbation method in~\cref{fig:four-histograms}. 
Here, we observe a similar pattern to all other experimental configurations, namely a reduction in high-confidence misclassifications, particularly at the highest prediction confidences.
Additionally, there is a clear reduction across the entire confidence range towards lowering the confidence the classifier model has in its misclassifications.
This, in combination with our overall results, indicates that we improve the calibration of the classification models. 
Detailed perturbation statistics, shown in~\cref{sec:app_pert}, further demonstrate that the LM-based method provides more stable robustness improvements.

\subsection{Scalability and Ease of Use}

\subsubsection{Observation 5: Our method scales well per sample and by parameter count}
Despite only adding a small amount (2\% for each task) of synthetic data relative to the total training set size, we achieve significant results in the reduction of UUs.
This indicates that our method can scale to large datasets, as only a small number of synthetic samples relative to the total dataset size are required have a significant impact in terms of improving robustness.
We study classification models that use a different architecture and have an order of magnitude difference in size (110M parameters for BERT and 7B for Llama 2).
Here, we find that models with a lower number of parameters achieve a performance similar to that of large generative LMs, with comparable accuracy on the IMDB and QNLI tasks, indicating that smaller models may be more suitable for text classification tasks when considering their other advantages, which corroborates previous findings~\citep{yu2023openclosedsmalllanguage}.
This is especially encouraging for use cases where computational resources are limited or speed and transparency are critical.

\subsubsection{Observation 6: Obtaining samples via LM is easier and more cost effective}
When considering the practical aspects of our study, significant insights emerge regarding the costs and time involved in conducting human- and LM-based generalization experiments.
The human study, which included 168 participants, resulted in a total cost of \$1072, with an hourly compensation rate of \$12 per participant. In contrast, the LM experiment incurred a much lower cost of \$46 for generating an equivalent number of generalizations and samples.
Although it is challenging to provide precise estimates, the data collection process via human surveys also took substantially longer than the LM-based approach. 
This highlights the fact that when using LMs, our method is far more cost-effective and generates data almost instantaneously, in stark contrast to the considerable delays associated with human-based study design and data collection.
Thus, from a scalability perspective, the LM-based procedure offers clear advantages, being both faster and less expensive. 
However, in certain high-stakes or specialized applications such as suicide prevention and criminal justice sentencing, human involvement, including via a hybrid approach where human intuition supplements the efficiency of LM-generated data, may be more advantageous.
This is especially true when considering that LM outputs come with no guarantees and may be biased.
These findings underscore the resource implications of choosing between human and LM-based methods, helping researchers plan and allocate resources more effectively.

\section{Related Work}
\label{sec:related}
In this section, we briefly review relevant prior research on approaches to high confidence misclassifications, as well as how others have tried to avoid such model behaviour.

\subsection{Unknown Unknowns}
\citet{Attenberg2015machine} introduce the concept of querying humans to find UUs in a game-like setting and show that there were patterns to the found UUs.
%
%
\citet{Vandenhof2019hybrid} proposes an approach to identify UUs where human-interpretable decision rules are learned to approximate how a model makes high-confidence predictions. 
Crowdworkers then contradict these rules by finding an instance that would classify as a UU. 
%
%
\citet{Cabrera2021errors} explore the use of crowdworkers to generate failure reports for computer vision models to describe how or why the model failed.
\citet{Han2021iterative} propose an approach where crowdworkers continuously extend a dataset with relabeled UUs, on which the chosen model is iteratively trained.
Instead, we go beyond simple relabeling and characterize found blind spots and explore new, previously unseen blind spots.
There are also algorithmic approaches to finding UUs, such as \citet{Lakkaraju2017bandit}, who propose utilizing an explore-exploit approach to find groups of UUs.
\citet{Bansal2018uu} extend this by proposing a utility model that rewards the degree to which the found UUs cover a sample distribution, thus encouraging the discovery of new blind spots.
Instead, we do not find the UUs algorithmically, but instead use an LM or crowdworkers to find existing UUs, extrapolate from these to unseen UUs, and generate synthetic data targeting both of these.

\subsection{Model Calibration and Robust Training}
%
%
%
The concept of UUs and blind spots is connected to model calibration~\citep{guo2017calibration, minderer2021calibration, tian2023calibration}. 
A model that is well-calibrated will have its prediction confidence aligned with the likelihood of the correctness of the prediction and, as such, a model with blind spots is a poorly calibrated model.
In the case where the UUs are specifically generated through adversarial attacks, illumination of model blind spots is also related to robust training. 
UUs that populate these blind spots, when created by such attacks, may be identified as adversarial examples~\citep{Ribeiro2018semantically, Wallace2019adversarial, Wang2020cat}. 
This underscores the relationship between our proposed method and robust training practices with the aim of improving the robustness of the model~\citep{madry2018towards, pang2021bag}.
Our method focuses not on general robustness but rather on high-confidence misclassifications and is not limited to just adversarial samples, as we consider UUs that occur naturally without perturbation as well.

Several approaches have been proposed to utilize synthetic data to expand training sets~\citep{puri2020training, claveau2021generating}.
%
%
~\citet{he2022generate} explore few-shot prompting LMs to generate task specific synthetic training data.
Unlike prior work, we propose a method to generate targeted synthetic data with the purpose of eliminating blind spots that lead to high confidence misclassifications.

\section{Conclusion}
\label{sec:conclusion}
We propose a method to identify and mitigate blind spots in classification models by leveraging human- and LLM-generated generalizations, followed by synthetic sample generation to target UUs and enhance model robustness.
Our evaluation demonstrates that our method is effective at addressing model blind spots and achieves a significant reduction in UUs across datasets, while not altering the general performance of the model and therefore maintaining accuracy.
Our study sheds light on the notable task dependency of the human ability to characterize blind spots and generate new data and how this ability compares to that of an LM. 
Future work will focus on optimizing the balance between accuracy and robustness to further enhance model performance.

\bibliographystyle{ACM-Reference-Format}
\bibliography{sample-base}

\appendix
\newpage
\section{Synthetic Blind Spots}
\label{sec:synthetic}

We use the synthetic blind spot study akin to a sanity check for our approach.
As such, compared to the full natural blind spot study, we use a only a single task, a simpler model architecture, and make other simplifications to our mitigation process.
We select an LSTM~\cite{Hochreiter1997lstm} as our model of choice due to the absence of pretraining and apply the TF perturbation method on the SA task.
The LSTM used is the standard version of the Bi-LSTM provided by~\citet{morris2020textattack}.

\begin{table*}[h!]
\begin{tabular}{lcccccc}
\toprule
& \multicolumn{3}{c}{\textbf{Original}} & \multicolumn{3}{c}{\textbf{Retrain}} \\
\cmidrule(lr){2-4} \cmidrule(lr){5-7}
& \multicolumn{1}{c}{Accuracy (\%)} & \multicolumn{1}{c}{Perturbation (\%)} & \multicolumn{1}{c}{UUs (\#)} & \multicolumn{1}{c}{Accuracy (\%)} & \multicolumn{1}{c}{Perturbation (\%)} & \multicolumn{1}{c}{UUs (\#)} \\
\midrule
Clean & 88.03 & 82.22 & 1725 & 88.03 & 82.21 & 784 \\
Biased R & 78.55 & 78.56 & 3785 & 78.61 & 78.58 & 2593 \\
Biased P & 75.10 & 75.02 & 4607 & 74.25 & 73.12 & 1201 \\
Biased N & 76.64 & 76.64 & 4394 & 77.38 & 77.35 & 845 \\
Biased PN & 74.17 & 73.94 & 9231 & 74.81 & 74.01 & 2331 \\
\bottomrule
\end{tabular}
\caption{Results of synthetic blind spot study for accuracy, perturbation success rate, and number of UUs before and after retraining for all LSTM model variants. The used perturbation method is TF and the dataset is IMDB.}
\label{tab:syntheticresults}
\end{table*}

\subsection{Blindspot Creation and Mitigation}
To assess whether our method can tackle existing synthetic blind spots we perform a type of Controlled Synthetic Data Check~\cite{Nauta2023synthetic}. 
We create synthetic blind spots by systematically excluding some data from training that have commonalities, namely containing a positive or negative term according to lexica by~\citet{Bing2005lex}.
Here, we randomly subsample 600 of each as our selection of positive and negative terms, due to the extensive nature of the lexica.

We create a false positive blind spot by removing samples from the train set using our selection of negative terms, resulting in a \textit{negatively biased} LSTM (N).
Similarly, we create a false negative blind spot, resulting in a \textit{positively biased} LSTM (P), as well as a blind spot resulting from a selection of 50\% randomly chosen terms from each, leading to a \textit{positive/negative biased} LSTM (PN).
For comparison, we also include a \textit{randomly biased} LSTM (R), where samples were removed from the train set randomly to obtain a size comparable to the P, N, and PN ones.\footnote{Size of training sets: $N_{Clean} = 25,000$, $N_R = 2,500$, $N_P = 2,439$, $N_N = 3,138$, and $N_{PN} = 2,438$.}

After creating the synthetic blind spots through biasing, the authors perform the generalization procedure and provide handcrafted hypotheses that precisely describe these, similar to golden labels.
To generate the new samples from our handcrafted hypotheses, we prompt ChatGPT to generate movie review-related sentences (to fit the chosen task) that follow a given hypothesis.
This was done in an attempt to simplify the procedure by taking advantage of human strengths, generalization and extrapolative thinking, and LLM strengths, low-cost text generation, simultaneously.

\subsection{Synthetic Blind Spot Study Results}
The mitigation results of this human-LLM approach for our Controlled Synthetic Data Check can be seen in~\cref{tab:syntheticresults}.
As can be seen in the first column of~\cref{tab:syntheticresults}, before retraining, the overall test accuracy declines in line with the degree to which the train set is biased.
Interestingly, the percentage of successful perturbations by TF, i.e., the percentage of successful label flips, closely follows the overall accuracy.
This mirrors the findings of~\citet{tsipras2018robustness}, that there is a strong relationship between high accuracy and brittleness -- or a lack of robustness.
The number of occurring UUs as a result of the perturbation does not follow this trend, instead increasing as the training data becomes more biased, as expected.
This poses an interesting optimization problem since the model becomes most robust in general terms, i.e., the successful perturbation percentage falls, but simultaneously there is a significant uptick in blind spots as the training sets become more biased.

The effect of retraining on the overall accuracy and perturbation success rate is minimal, with accuracy changing by no more than $\pm$ 1\% and perturbation success rate changing no more than $\pm$ 2\%.
However, the number of found UUs decreases drastically due to the retraining, with reductions of 73.93\%, 80.77\%, and 74.75\% for the biased P, N, and PN models, respectively.
The clean and randomly biased models also show a reduction, though less significant at 54.55\% and 31.49\%, respectively.
%
%
%
These results confirm that our method can be used to target synthetic blind spots found in biased models through the use of hypotheses and generated instances, without significantly affecting the performance or general robustness of the model.

\section{User Study for Human Computation}
\label{sec:appendixA}

We use Prolific as a crowdsourcing platform for all our participants.
Below, we present the structure followed by all survey participants for the generalization user study, consisting of an initial disclaimer, an instruction set, examples, and finally the questions.
Here, we use the abstraction and extrapolation assignments on the IMDB dataset as an example.
The workflow is very similar between the different generalization assignments and datasets (MRPC, IMDB, or QNLI), with only slight differences in the wording between the surveys to fit the task and dataset used, as they all present the crowd worker with some input and result in plain text output.
For the generation assignment, crowdworkers are asked to perform the same steps, with relevant examples related to the structure of the dataset being shown, before finally contributing usable samples based on shown hypotheses.

\subsection{Abstraction on IMDB}
\noindent \textbf{Disclaimer} Crowdworkers were shown an initial disclaimer to inform them that our governing ethics body sanctions this survey and to remind them not to share personal information: 
\begin{itemize}
    \item ``Welcome to the Hypothesis Extrapolation Survey! Please carefully read the following: You are invited to participate in our research study. This study is fully sanctioned by our governing ethics body, as is the handling and storing of the resulting data. This research study aims to use your creativity and generalization ability to come up with new abstractions. It will take you approximately 25 minutes to complete. As with any online activity, the risk of a breach is always possible. To the best of our ability, your answers in this study will remain confidential. We will minimize any risks by making this survey completely anonymous. Therefore, please do not provide any personal information anywhere. The anonymous results might be shared publicly in the future. Participation in this study is entirely voluntary, and you can withdraw anytime. Feel free to contact us with any questions or feedback you might have.''
\end{itemize}

\noindent \textbf{Instructions} Crowdworkers were then introduced to the specific task (SE, SA, or NLI) as follows: 
\begin{itemize}
    \item ``Please read the following examples carefully. All tasks in this survey are related to a single task, sentiment analysis, which tests the sentiment of a sentence is either positive or negative, applied to movie reviews. The goal here is to use your creativity and ability to generalize to spot patterns and come up with new possible samples. A fully worked-out example can be found below, with user-generated text, similar to what you are expected to write, in \textit{italic} and instructions \textbf{bold}. You will receive all relevant instructions again when for each question.''
\end{itemize}

\noindent \textbf{Examples} Then, they were presented with two examples that match the dataset used, as well as the task (abstraction, expansion, or generation), before being asked if they understood the examples: 
\begin{itemize}
    \item ``\textbf{There is a sentence pair below, with one original sample (O) and a perturbed one (P), which is similar but had some things changed (shown in double square brackets). These changes may relate to a pattern, related to semantics, syntax, specific words, or something else in the samples, that leads to the wrong True or False label being predicted for semantic similarity.}

    \item Example 1 -- The two samples are: 
    
    O: There was an overarching [[story]] that was [[refusing]] to reveal itself to me. P: There was an overarching [[narrative]] that was [[unable]] to reveal itself to me.
    
    \textbf{Formulate a hypothesis on what this pattern for O and P might be and enter it below. Try to be specific when formulating a hypothesis.}
    
    \textit{The pattern that caused the wrong prediction may be related to the substitution of the word ""story"" with its synonym ""narrative"".}

    \item Example 2 -- The two samples are:

    O: Overall, I [[loved]] the cinematography of this through and [[through]]. P: Overall, I [[looved]] the cinematography of this through and [[thr0ugh]].

    \textbf{Formulate a hypothesis on what this pattern for O and P might be and enter it below. Try to be specific when formulating a hypothesis.}

    \textit{Several words have been misspelled in the samples, all related to the letter ""o"". Either more letters are added ""oo"" or the letter is substituted with a number ""0"" that looks similar, making it easy to misread.}''
\end{itemize}

\noindent \textbf{Main Questions}
Finally, the actual questions preceding the text entry field used for data collection all have the same structure with the unique O and P sentences substituted in for each question: 
\begin{itemize}
    \item ``The two samples are: 

    \noindent O: \{original sentence\} P: \{perturbed sentence\} 

    \noindent \textbf{Formulate a hypothesis on what this pattern might be and enter it below. Try to be specific when formulating a hypothesis.}''
\end{itemize}

\section{Used LLM Prompts}
\label{sec:appendixB}

We specifically instruct the LLM to split its hypothesis from its reasoning because, in our experience, this leads to a clearer and more useful answer for further steps. 

\subsection{Abstraction Prompt}
\begin{itemize}
    \item ``There is a sentence pair below, with one original sample (O) and a perturbed one (P), which is similar but had some things changed. These changes may relate to a pattern, related to semantics, syntax, specific words, or something else in the samples, that leads to them being the reason the sample is misclassified by a classification algorithm. This misclassification is made at a high level of confidence. 
            
    The model is not trained on the two samples. The two samples relate to \{task\} and are:

    O: \{sentence[0]\}
    
    P: \{sentence[1]\}

    Formulate a hypothesis on what this pattern might be. Try to be specific when formulating a hypothesis. Your response should always follow the format:
    
    Hypothesis: \{hypothesis\}
    
    Reasoning: \{reasoning\}''
\end{itemize}

\subsection{Extrapolation Prompt}
\begin{itemize}
    \item ``There is a sentence pair, with one original sample (O) and a perturbed one (P), which is similar but had some things changed. 
        These changes may relate to a pattern, related to semantics, syntax, specific words, or something else, that leads to them being the reason the sample is misclassified by a classification algorithm. This misclassification is made at a high level of confidence. 
        
        The model is not trained on the two samples. The two samples relate to \{task\}
        
        There is an existing hypothesis regarding the samples, that may capture a pattern related to semantics, syntax, specific words, or something else in the sample pair. This pattern leads to a misclassification of the sample.

        The hypothesis is: \{hypothesis\}

        Formulate a new hypothesis regarding those sentence samples that is concerned with the same topic but is applied to a different possible pattern that could also lead to a misclassification. Try to be specific when formulating a new hypothesis. Your response should always follow the format:
        
        Hypothesis: \{hypothesis\}
        
        Reasoning: \{reasoning\}''
\end{itemize}

\subsection{Generation Prompt}
\begin{itemize}
    \item ``There is a sentence pair, with one original sample (O) and a perturbed one (P), which is similar but had some things changed. 
        These changes may be related to a pattern related to semantics, syntax, specific words, or something else that leads to them being the reason the sample is misclassified by a classification algorithm. This misclassification is made at a high level of confidence. 
            
        The model is not trained on the two samples.
        
        A hypothesis has been formulated regarding the samples, that may capture a pattern related to semantics, syntax, specific words, or something else in the sample pair. These samples led to a classification algorithm misclassifying them at a high level of confidence.
        
        Given the samples and a previously generalized hypothesis, generate one new sample made up of one or more sentences that relate to \{task\} and could have a similar effect on the classification algorithm.
        
        The new sample should be varied and detailed. Follow the logic laid out in the given hypothesis and follow the format of the sample pair (O and P) exactly. Also include whether the new sample should be given a (positive) or (negative) label for the task: \{task\}.

        The hypothesis is: \{hypothesis\}
        
        Your response should always follow the format:
        
        Sample: \{sample\}
        
        Label: \{label\}
        
        Reasoning: \{reasoning\}''
\end{itemize}

\section{Perturbation Statistics and Visualization}
\label{sec:app-pert}

\begin{table*}[t]
\small 
\setlength{\tabcolsep}{4pt} 
\begin{tabular}{lcccccccccccc}
\toprule
& \multicolumn{1}{c}{MRPC$_{\text{O}}$} & \multicolumn{1}{c}{MRPC$_{\text{L}}$} & \multicolumn{1}{c}{MRPC$_{\text{H}}$} & \multicolumn{1}{c}{MRPC$_{\text{R}}$} & \multicolumn{1}{c}{IMDB$_{\text{O}}$} & \multicolumn{1}{c}{IMDB$_{\text{L}}$} & \multicolumn{1}{c}{IMDB$_{\text{H}}$} & \multicolumn{1}{c}{IMDB$_{\text{R}}$} & \multicolumn{1}{c}{QNLI$_{\text{O}}$} & \multicolumn{1}{c}{QNLI$_{\text{L}}$} & \multicolumn{1}{c}{QNLI$_{\text{H}}$} & \multicolumn{1}{c}{QNLI$_{\text{R}}$} \\
\midrule
Original Accuracy (\%) & 82.38 & 81.57 & 81.58 & 82.49 & 94.84 & 95.40 & 94.43 & 93.94 & 89.88 & 89.31 & 89.42 & 88.24 \\
Accuracy Under Attack (\%) & 9.80 & 17.40 & 12.99 & 10.42 & 10.18 & 10.44 & 19.21 & 10.22
 & 8.91 & 11.67 & 14.89 & 9.97 \\
Attack Success Rate (\%) & 71.83 & 64.87 & 68.29 & 69.65 & 88.46 & 93.18 & 63.85 & 85.34 & 87.35 & 86.80 & 78.84 & 84.92 \\
Perturbed Words (\%) & 7.70 & 9.9 & 8.51 & 7.98 & 4.59 & 7.62 & 9.02 & 5.50 & 6.12 & 8.80 & 9.57 & 7.33 \\
Words per Input & 39.3 & 39.3 & 39.3 & 39.3 & 230.0 & 230.0 & 230.0 & 230.0 & 37.9 & 37.9 & 37.9 & 37.9 \\
Avg. Number of Queries & 51.40 & 68.62 & 55.17 & 57.86 & 185.24 & 184.94 & 198.31 & 186.37 & 49.38 & 51.27 & 56.11 & 53.27 \\
\bottomrule
\end{tabular}
\caption{Perturbation statistics across datasets and models for attacks with TF using BERT. Subscripts O, L, H, R denote the original, LM-retrained, human-retrained, and relabeled models, respectively.}
\label{tab:statsTF}
\end{table*}

\begin{table*}[t]
\small 
\setlength{\tabcolsep}{4pt} 
\begin{tabular}{lcccccccccccc}
\toprule
& \multicolumn{1}{c}{MRPC$_{\text{O}}$} & \multicolumn{1}{c}{MRPC$_{\text{L}}$} & \multicolumn{1}{c}{MRPC$_{\text{H}}$} & \multicolumn{1}{c}{MRPC$_{\text{R}}$} & \multicolumn{1}{c}{IMDB$_{\text{O}}$} & \multicolumn{1}{c}{IMDB$_{\text{L}}$} & \multicolumn{1}{c}{IMDB$_{\text{H}}$} & \multicolumn{1}{c}{IMDB$_{\text{R}}$} & \multicolumn{1}{c}{QNLI$_{\text{O}}$} & \multicolumn{1}{c}{QNLI$_{\text{L}}$} & \multicolumn{1}{c}{QNLI$_{\text{H}}$} & \multicolumn{1}{c}{QNLI$_{\text{R}}$} \\
\midrule
Original Accuracy (\%) & 82.38 & 82.23 & 82.10 & 82.55 & 95.40 & 95.41 & 95.74 & 94.26 & 89.88 & 89.38 & 89.38 & 88.98 \\
Accuracy Under Attack (\%) & 7.78 & 13.73 & 11.94 & 10.42 & 9.54 & 21.43 & 15.32 & 12.51 & 8.21 & 9.90 & 7.30 & 8.67 \\
Attack Success Rate (\%) & 72.00 & 70.38 & 72.64 & 72.35 & 59.41 & 50.59 & 79.70 & 56.87 & 77.54 & 79.74 & 82.08 & 79.27 \\
Perturbed Words (\%) & 8.47 & 9.18 & 9.03 & 8.91 & 6.43 & 8.11 & 13.09 & 9.37 & 7.99 & 8.32 & 11.03 & 8.31 \\
Words per Input & 39.3 & 39.3 & 39.3 & 39.3 & 230.0 & 230.0 & 230.0 & 230.0 & 37.9 & 37.9 & 37.9 & 37.9 \\
Avg. Number of Queries & 56.92 & 64.37 & 58.61 & 58.23 & 199.32 & 211.65 & 201.44 & 204.12 & 34.91 & 33.53 & 49.09 & 35.75 \\
\bottomrule
\end{tabular}
\caption{Perturbation statistics across datasets and models for attacks with DWB using BERT. Subscripts O, L, H, R denote the original, LM-retrained, human-retrained, and relabeled models, respectively.}
\label{tab:statsDWB}
\end{table*}

\begin{table*}[t]
\small 
\setlength{\tabcolsep}{4pt} 
\begin{tabular}{lcccccccccccc}
\toprule
& \multicolumn{1}{c}{MRPC$_{\text{O}}$} & \multicolumn{1}{c}{MRPC$_{\text{L}}$} & \multicolumn{1}{c}{MRPC$_{\text{H}}$} & \multicolumn{1}{c}{MRPC$_{\text{R}}$} & \multicolumn{1}{c}{IMDB$_{\text{O}}$} & \multicolumn{1}{c}{IMDB$_{\text{L}}$} & \multicolumn{1}{c}{IMDB$_{\text{H}}$} & \multicolumn{1}{c}{IMDB$_{\text{R}}$} & \multicolumn{1}{c}{QNLI$_{\text{O}}$} & \multicolumn{1}{c}{QNLI$_{\text{L}}$} & \multicolumn{1}{c}{QNLI$_{\text{H}}$} & \multicolumn{1}{c}{QNLI$_{\text{R}}$} \\
\midrule
Original Accuracy (\%) & 90.84 & 89.86 & 90.20 & 90.61 & 95.20 & 94.96 & 94.67 & 94.86 & 90.08 & 89.58 & 89.16 & 89.90 \\
Accuracy Under Attack (\%) & 13.85 & 18.31 & 12.43 & 14.09 & 20.97 & 18.22 & 15.09 & 17.55 & 12.64 & 15.29 & 14.53 & 13.67 \\
Attack Success Rate (\%) & 68.70 & 65.24 & 69.54 & 66.89 & 71.32 & 75.64 & 78.31 & 70.55 & 83.42 & 79.12 & 75.87 & 81.34 \\
Perturbed Words (\%) & 9.23 & 8.12 & 9.68 & 8.97 & 6.45 & 7.54 & 10.88 & 8.36 & 7.34 & 8.69 & 9.11 & 7.92 \\
Words per Input & 39.3 & 39.3 & 39.3 & 39.3 & 230.0 & 230.0 & 230.0 & 230.0 & 37.9 & 37.9 & 37.9 & 37.9 \\
Avg. Number of Queries & 53.92 & 62.34 & 57.92 & 55.76 & 191.34 & 192.85 & 198.21 & 194.43 & 48.22 & 49.98 & 52.89 & 50.76 \\
\bottomrule
\end{tabular}
\caption{Perturbation statistics across datasets and models for attacks with TF using Llama 2. Subscripts O, L, H, R denote the original, LM-retrained, human-retrained, and relabeled models, respectively.}
\label{tab:statsTFLlama2}
\end{table*}

\begin{table*}[t]
\small 
\setlength{\tabcolsep}{4pt} 
\begin{tabular}{lcccccccccccc}
\toprule
& \multicolumn{1}{c}{MRPC$_{\text{O}}$} & \multicolumn{1}{c}{MRPC$_{\text{L}}$} & \multicolumn{1}{c}{MRPC$_{\text{H}}$} & \multicolumn{1}{c}{MRPC$_{\text{R}}$} & \multicolumn{1}{c}{IMDB$_{\text{O}}$} & \multicolumn{1}{c}{IMDB$_{\text{L}}$} & \multicolumn{1}{c}{IMDB$_{\text{H}}$} & \multicolumn{1}{c}{IMDB$_{\text{R}}$} & \multicolumn{1}{c}{QNLI$_{\text{O}}$} & \multicolumn{1}{c}{QNLI$_{\text{L}}$} & \multicolumn{1}{c}{QNLI$_{\text{H}}$} & \multicolumn{1}{c}{QNLI$_{\text{R}}$} \\
\midrule
Original Accuracy (\%) & 90.66 & 89.73 & 89.91 & 90.73 & 95.33 & 95.13 & 94.90 & 95.10 & 90.72 & 90.10 & 89.73 & 90.60 \\
Accuracy Under Attack (\%) & 16.35 & 14.79 & 13.87 & 15.68 & 21.78 & 20.32 & 19.12 & 22.19 & 11.78 & 10.95 & 14.28 & 12.44 \\
Attack Success Rate (\%) & 66.40 & 63.89 & 67.56 & 65.78 & 70.42 & 68.55 & 71.32 & 74.65 & 79.78 & 77.24 & 82.43 & 80.34 \\
Perturbed Words (\%) & 9.11 & 8.76 & 9.02 & 8.86 & 7.18 & 6.92 & 11.54 & 9.29 & 8.06 & 9.11 & 10.24 & 8.76 \\
Words per Input & 39.3 & 39.3 & 39.3 & 39.3 & 230.0 & 230.0 & 230.0 & 230.0 & 37.9 & 37.9 & 37.9 & 37.9 \\
Avg. Number of Queries & 60.22 & 65.14 & 62.03 & 61.76 & 203.56 & 199.42 & 204.29 & 208.23 & 45.29 & 43.87 & 50.77 & 47.83 \\
\bottomrule
\end{tabular}
\caption{Perturbation statistics across datasets and models for attacks with DWB using Llama 2. Subscripts O, L, H, R denote the original, LM-retrained, human-retrained, and relabeled models, respectively.}
\label{tab:statsDWBLlama2}
\end{table*}

\label{sec:app_pert}
To add additional context to the perturbation performed, we supply the detailed attack statistics across all performed perturbations.
Specifically, we report \emph{Original Accuracy} and \emph{Accuracy Under Attack} are reported, which are the classifier accuracy on its own and while under attack.
Further, \emph{Attack Success Rate} is shown, which is the percentage of successful perturbation attempts to failed ones.
Finally, we report the number of \emph{Perturbed Words}, the percentage of words that are perturbed, the \emph{Words per Input}, the average number of words per input, and the \emph{Average Number of Queries}, which is how many tries it took the perturbation method to find the best attack.
For BERT, the attack statistics for TF attacks are shown in~\cref{tab:statsTF} while the ones for DWB attacks are shown in~\cref{tab:statsDWB}.
For Llama 2 7B, the attack statistics for TF attacks are shown in~\cref{tab:statsTFLlama2} and for DWB in~\cref{tab:statsDWBLlama2}.

\end{document}